\PassOptionsToPackage{dvipsnames,table}{xcolor}

\documentclass[sigconf]{acmart}
\usepackage[dvipsnames,table]{xcolor}
\usepackage{booktabs} 
\usepackage{tabularx}

\usepackage{multirow}
\usepackage{multicol}
\usepackage{makecell}
\usepackage{balance}
\usepackage[utf8]{inputenc}
\usepackage{kotex}
\usepackage{enumitem}
\usepackage[ruled,lined, linesnumbered]{algorithm2e}
\usepackage{algpseudocode}
\usepackage{threeparttable}
\usepackage{array}
\usepackage{tikz}
\usetikzlibrary{calc}
\usepackage{ragged2e}
\usepackage{svg}

\DeclareMathOperator*{\argmin}{arg\,min}

\newcommand{\red}[1]{\cellcolor{Maroon!8}\textcolor{BrickRed!30!black}{#1}}
\newcommand{\blue}[1]{\cellcolor{SeaGreen!8}\textcolor{RoyalBlue!30!black}{#1}}
\newcommand{\purple}[1]{\cellcolor{Orchid!8}\textcolor{Plum!30!black}{#1}}

\newcommand{\textred}[1]{\colorbox{Maroon!8}{\textcolor{BrickRed!30!black}{#1}}}
\newcommand{\textblue}[1]{\colorbox{SeaGreen!8}{\textcolor{RoyalBlue!30!black}{#1}}}
\newcommand{\textpurple}[1]{\colorbox{Orchid!8}{\textcolor{Plum!30!black}{#1}}}

\newlength\mylen

\usepackage{tabto}

\let\oldnl\nl
\newcommand{\nonl}{\renewcommand{\nl}{\let\nl\oldnl}}
\newcolumntype{P}[1]{>{\centering\arraybackslash}p{#1}}

\setcopyright{acmcopyright}

\acmDOI{10.1145/3672608.3707957}

\acmISBN{978-1-4503-8713-2/22/04}

\acmConference[SAC'25]{ACM SAC Conference}{March 31 - April 4, 2025}{Sicily, Italy}  
\acmYear{2025}
\copyrightyear{2025}

\begin{document}
\title[]{UKTA: Unified Korean Text Analyzer}

\author{Seokho Ahn}
\authornote{ Equal contribution.}
\orcid{0000-0002-5715-4057}
\affiliation{%
  \institution{Inha University}
  \city{Incheon} 
  \state{South Korea} 
}
\email{sokho0514@inha.edu}

\author{Junhyung Park}
\authornotemark[1]
\orcid{0009-0000-6714-9684}
\affiliation{%
  \institution{Inha University}
  \city{Incheon} 
  \state{South Korea} 
}
\email{quixote1103@inha.edu}

\author{Ganghee Go}
\authornotemark[1]
\orcid{0009-0006-1027-105X}
\affiliation{%
  \institution{Inha University}
  \city{Incheon} 
  \state{South Korea} 
}
\email{khko99@inha.edu}

\author{Chulhui Kim}
\orcid{0000-0003-4161-2882}
\affiliation{%
  \institution{Inha University}
  \city{Incheon} 
  \state{South Korea} 
}
\email{clearfe@inha.ac.kr}

\author{Jiho Jung}
\orcid{0009-0000-5711-1834}
\affiliation{%
  \institution{Inha University}
  \city{Incheon} 
  \state{South Korea} 
}
\email{jacob\_jjh@inha.ac.kr}

\author{Myung Sun Shin}
\orcid{0009-0003-1832-8987}
\affiliation{%
  \institution{Inha University}
  \city{Incheon} 
  \state{South Korea} 
}
\email{rescript@inha.ac.kr}

\author{Do-Guk Kim}
\authornotemark[2]
\orcid{0000-0002-3564-7002}
\affiliation{%
  \institution{Inha University}
  \city{Incheon} 
  \state{South Korea} 
}
\email{dgkim@inha.ac.kr}

\author{Young-Duk Seo}
\authornote{ Co-corresponding authors.}
\orcid{0000-0001-8542-2058}
\affiliation{%
  \institution{Inha University}
  \city{Incheon} 
  \state{South Korea} 
}
\email{mysid88@inha.ac.kr}

\renewcommand{\shortauthors}{Ahn et al.}

\begin{abstract}
Evaluating writing quality is complex and time-consuming often delaying feedback to learners. While automated writing evaluation tools are effective for English, Korean automated writing evaluation tools face challenges due to their inability to address multi-view analysis, error propagation, and evaluation explainability. To overcome these challenges, we introduce \textsf{UKTA} (\textbf{U}nified \textbf{K}orean \textbf{T}ext \textbf{A}nalyzer), a comprehensive Korea text analysis and writing evaluation system. \textsf{UKTA} provides accurate low-level morpheme analysis, key lexical features for mid-level explainability, and transparent high-level rubric-based writing scores. Our approach enhances accuracy and quadratic weighted kappa over existing baseline, positioning \textsf{UKTA} as a leading multi-perspective tool for Korean text analysis and writing evaluation.
\end{abstract}

%
%



\begin{teaserfigure}
  \includegraphics[width=0.95\textwidth]{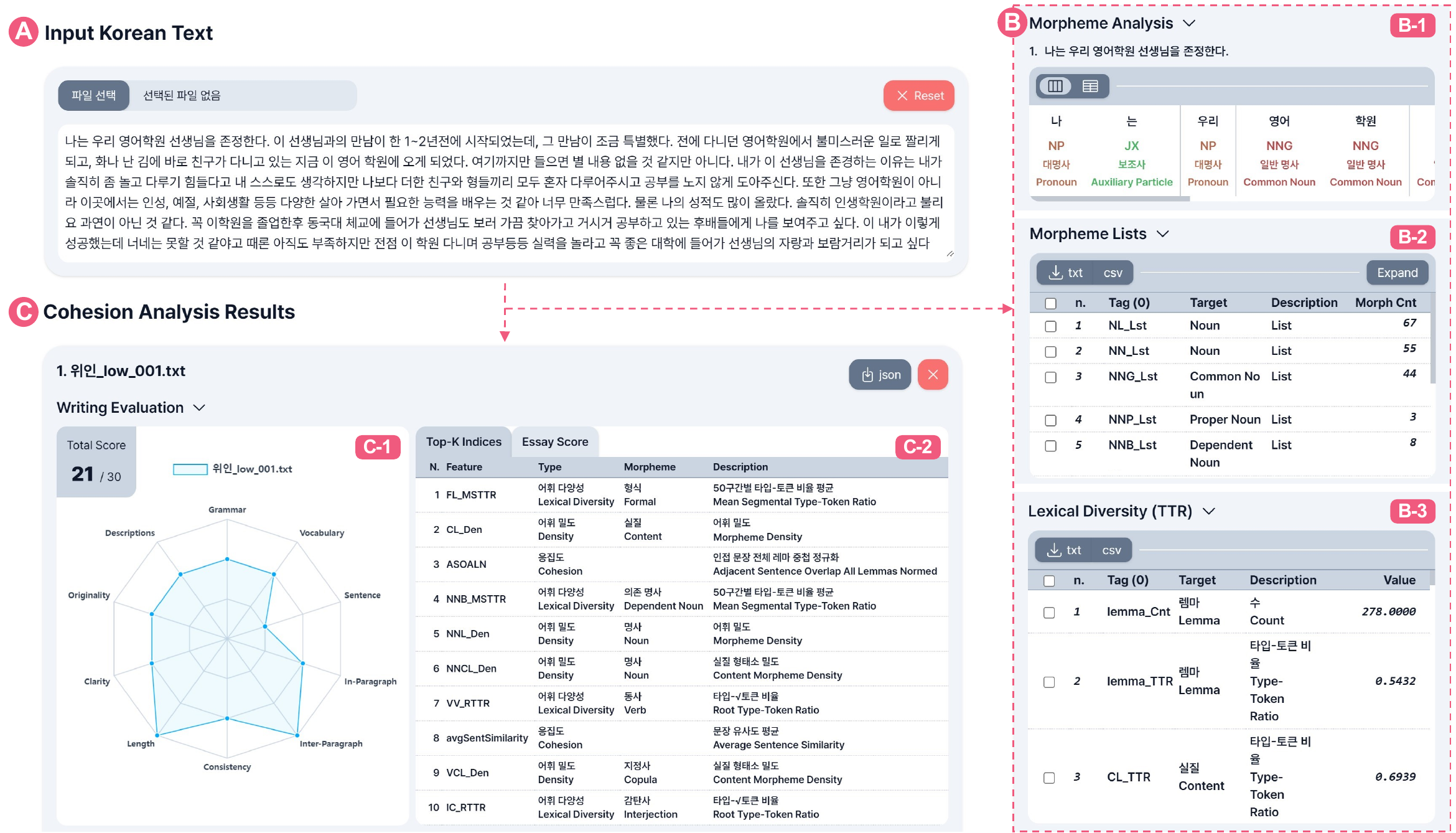}
  \centering
  \caption{\textsf{UKTA} is a comprehensive Korean text analyzer that provides morpheme analysis, lexical feature analysis, and explainable writing evaluation: \normalfont{\textbf{(A)} Users can input Korean text as a file or paragraph, \textbf{(B)} Display multi-perspective results such as morphemes and lexical features, and \textbf{(C)} Provide explainable, visualized writing evaluation results in the form of rubric scores, along with the top features that contributed to the scores. Users can download these results in various formats, including JSON, TXT, and CSV files.}}
  \label{fig:teaser}
  \vspace{\baselineskip}
\end{teaserfigure}

\maketitle

\section{Introduction}
\label{sec:01_introduction}

\textit{Writing matters}, but reaching a consensus on writing quality standards can be challenging \cite{deane2022importance}. 
Writing evaluation is a complex task that requires significant time and effort from professional evaluators, making it difficult to provide timely feedback to students \cite{finn2024memory}.
To address this issue, various English text analyzers \cite{graesser2004cohmetrix, mcnamara2010linguistic, rei2016sentence, crossley2019taaco} and automated writing evaluation tools \cite{wang2022aessota, jeon2021countering, uto2020neural} have been developed. 
Recent automated writing evaluation systems can produce scores that align closely with human evaluation in certain contexts \cite{beigman2020automated}. 
This success has led to their integration into standardized English tests such as the Test of English as a Foreign Language (TOEFL\footnote{https://www.ets.org/toefl.html}) and the Graduate Record Examination (GRE\footnote{https://www.ets.org/gre.html}), providing both automated and human evaluation scores \cite{beigman2020automated, ramineni2012gre}.

Building on the success of automated English writing evaluation systems, we examine three key factors necessary for practical and widespread use of Korean text analysis and writing evaluation: 
\textit{\textsf{(i) Multi-view analysis.}} Automated writing evaluation should consider multiple perspectives, from low-level (\textit{i.e.}, concrete) analyses such as morpheme analysis and lexical diversity
, to higher-level (\textit{i.e.}, abstract) analyses such as semantic cohesion and automatic writing evaluation; 
\textit{\textsf{(ii) Error propagation.}} Errors occurring in early stages (\textit{e.g.,} morpheme analysis) should have minimal impact in later stages (\textit{e.g.,} writing evaluation). This is particularly important in Korean, an \textit{agglutinative} language, where frequent morphological changes make it more vulnerable to error propagation \cite{matteson2018rich};
and \textit{\textsf{(iii) Evaluation explainability.}} High-level, abstract evaluation results should be interpretable by humans, who need to understand the reason behind the scores and the features that influenced the results. 
Providing this explainability to users is crucial for ensuring reliability, as these tools have the potential to make mistakes; Unfortunately, existing Korean text analyzers \cite{ryu2019koranlysis, lee2024exploring, kim2024korcat} and automated writing evaluation tools \cite{lee2022argument, lee2023pasta} do not fully meet all these requirements, limiting their practical use.

To address the research gap, we introduce \textsf{UKTA} (\textbf{U}nified \textbf{K}orean \textbf{T}ext \textbf{A}nalyzer), a comprehensive Korean text analysis system for evaluating Korean writing. 
First, we provide accurate low-level analysis based on 
state-of-the-art Korean morpheme
analyzer, which minimizes error propagation. 
In addition to morpheme analysis, we categorize and provide key features, such as lexical richness and semantic cohesion, at the mid-level to enable explainable writing evaluation.
Finally, we present a comprehensive rubric-based writing score as a high-level metric based on a novel attention-based deep learning method and provide the features contributing to that score to enhance explainability and reliability. 
Notably, using all the suggested features improves writing evaluation performance compared to baseline in terms of accuracy and quadratic weighted kappa scores.
To the best of our knowledge, \textsf{UKTA} is the first comprehensive Korean text analysis and writing evaluation tool providing accurate results from a multi-view perspective.

Our contributions are summarized as follows:

\begin{itemize}[leftmargin=1.1em]
\item We introduce \textsf{UKTA}, a comprehensive Korean text analysis and writing evaluation system from multiple perspectives, and present a tool for its practical application.

\item Rather than simply presenting writing evaluation scores, \textsf{UKTA} enhances explainability and reliability by providing detailed feature scores such as morpheme, lexical diversity, and cohesion.

\item Experimental results demonstrate that writing evaluation accuracy improves when the proposed features are considered, compared to scores derived solely from raw text.
\end{itemize} 

The remainder of this paper is organized as follows.
Section \ref{sec:02_relatedworks} reviews the related work.
Section \ref{sec:04_model} introduces the proposed approach in detail.
In Section \ref{sec:05_experiments}, we present the experimental results. Finally, Section \ref{sec:06_conclusion} concludes our work and outlines future work.

\section{Related Work}
\label{sec:02_relatedworks}

\subsection{Text analyzers} 

Text analysis has become an essential tool for evaluating written content, particularly in language education, and linguistic research \cite{mcnamara2010linguistic, ha2019lexical}. Modern English text analyzers utilize a combination of lexical and semantic metrics to assess the quality, coherence, and complexity of a text \cite{crossley2016tool}. These tools break down a text into measurable features, such as lexical diversity and cohesion, to provide quantitative insights. Although English text analyzers have seen significant success in both academic and practical applications, applying similar methodologies to Korean text analysis has been challenging due to linguistic differences. This section reviews key works on lexical diversity and cohesion, which are fundamental components in text analysis systems. 

Lexical density is a key indicator of a writer's vocabulary depth and is important for evaluating text quality in English. Common measures include the number of different words (NDW) \cite{miller1991contextual} and type/token ratios (TTR, RTTR, CTTR) \cite{chotlos1944iv, guiraud1959problemes, carroll1964language} to assess lexical diversity \cite{van2007comparing}. Advanced metrics like MSTTR \cite{johnson1944studies}, MATTR \cite{covington2007mattr}, MTLD \cite{mccarthy2005assessment}, HD-D \cite{mccarthy2010mtld}, and vocd-D \cite{mccarthy2007vocd} provide more detailed evaluations by analyzing text in fixed lengths or by considering vocabulary complexity. These measures strongly correlate with writing quality and lexical diversity \cite{ha2019lexical}. They are crucial in language assessment, where they complement human evaluations by offering objective, efficient assessments. Meanwhile, studies on Korean text analysis have been less extensive due to the difficulty of automated morpheme analysis. Although there have been attempts to apply lexical diversity measures to Korean \cite{lee2024exploring}, no comprehensive system has been developed that is easily accessible for automated evaluation or for use by educators for further analysis. This gap highlights the need for more robust and user-friendly tools to facilitate deeper exploration of Korean text analysis. 

Language models have emerged as prominent tools for evaluating text, offering sophisticated methods to assess various linguistic features such as coherence, complexity, and cohesion. Semantic cohesion, in particular, evaluates the consistency of a topic within a paragraph (topic consistency) and the similarity of meanings across sentences (sentence similarity). Transformer-based models, like BERT and SBERT, have been effectively utilized for measuring semantic cohesion in English texts \cite{doewes2022individual, rei2016sentence, ramesh2022automated}. However, despite the success of these models in English, there has been limited adoption of transformer-based language models for semantic cohesion analysis in Korean text. The unique linguistic characteristics of Korean, along with the challenges of morpheme segmentation, have slowed the development of such systems.

\subsection{Automated Writing Evaluation Tools}

Recently, the field of Natural Language Processing (NLP) has advanced significantly, leading to an increased demand for automated writing evaluation systems and prompting extensive research in this area \cite{jeon2021countering, uto2020neural, wang2022aessota}. The development of transformer-based models, such as BERT, has been particularly significant for automated writing evaluation, representing a breakthrough in the field. \cite{wang2022aessota} leverage this cutting-edge technology to develop an automated writing evaluation model that generates multi-scale essay representation vectors. Specifically, this model utilizes BERT’s powerful sentence representation and essay learning capabilities to evaluate multiple aspects of essays. This research achieves state-of-the-art performance in the Automated Student Assessment Prize (ASAP)\footnote{https://paperswithcode.com/dataset/asap} task, which is based on English writing evaluation datasets.

In the context of automated writing evaluation for Korean, several studies have also been actively conducted. Notably, the National Information Society Agency (NIA) established the Korean \textsf{Essay Evaluation Dataset}\footnote{https://url.kr/qilx39} in 2021 through its \textsf{AI-HUB}\footnote{https://www.aihub.or.kr/} platform, providing a crucial resource for research on automated writing evaluation tools for Korean. In addition to supplying the dataset, AI-HUB introduced a baseline evaluation model, which has since served as a starting point for further development of Korean automated evaluation systems. For instance, \cite{lee2022argument} proposes an automated writing evaluation model that maximizes the potential of the Korean dataset by combining argument mining techniques and a RoBERTa-based model pre-trained on the Korean Language Understanding Evaluation (KLUE) \cite{park2021klue} dataset. Their model effectively analyzes the logical structure of Korean essays by generating representation vectors that accurately reflect argumentative structures. Moreover, \cite{lee2023pasta} suggests \textsf{PASTA-I}, a KoELECTRA \cite{kim2020lmkor}-based automated scoring system for Korean essays and written responses, utilizing the \textsf{Essay Evaluation Dataset}.

However, these models relied on sentence-piece tokenizers, which are primarily designed for English, rather than Korean-specific morpheme-based tokenizers suited to the complex agglutinative structure of the Korean language. This makes the models more susceptible to error propagation when analyzing Korean. Additionally, these models did not provide multi-view analysis or sufficient explanation regarding the evaluation process, which makes it challenging to ensure the reliability of the evaluations, particularly given the inherent complexity of the language. As a result, these models have faced difficulties in delivering accurate and trustworthy assessments of Korean essays.

\section{Unified Korean Text Analyzer}
\label{sec:04_model}
\begin{figure}[t]
\includegraphics[width=1\columnwidth]{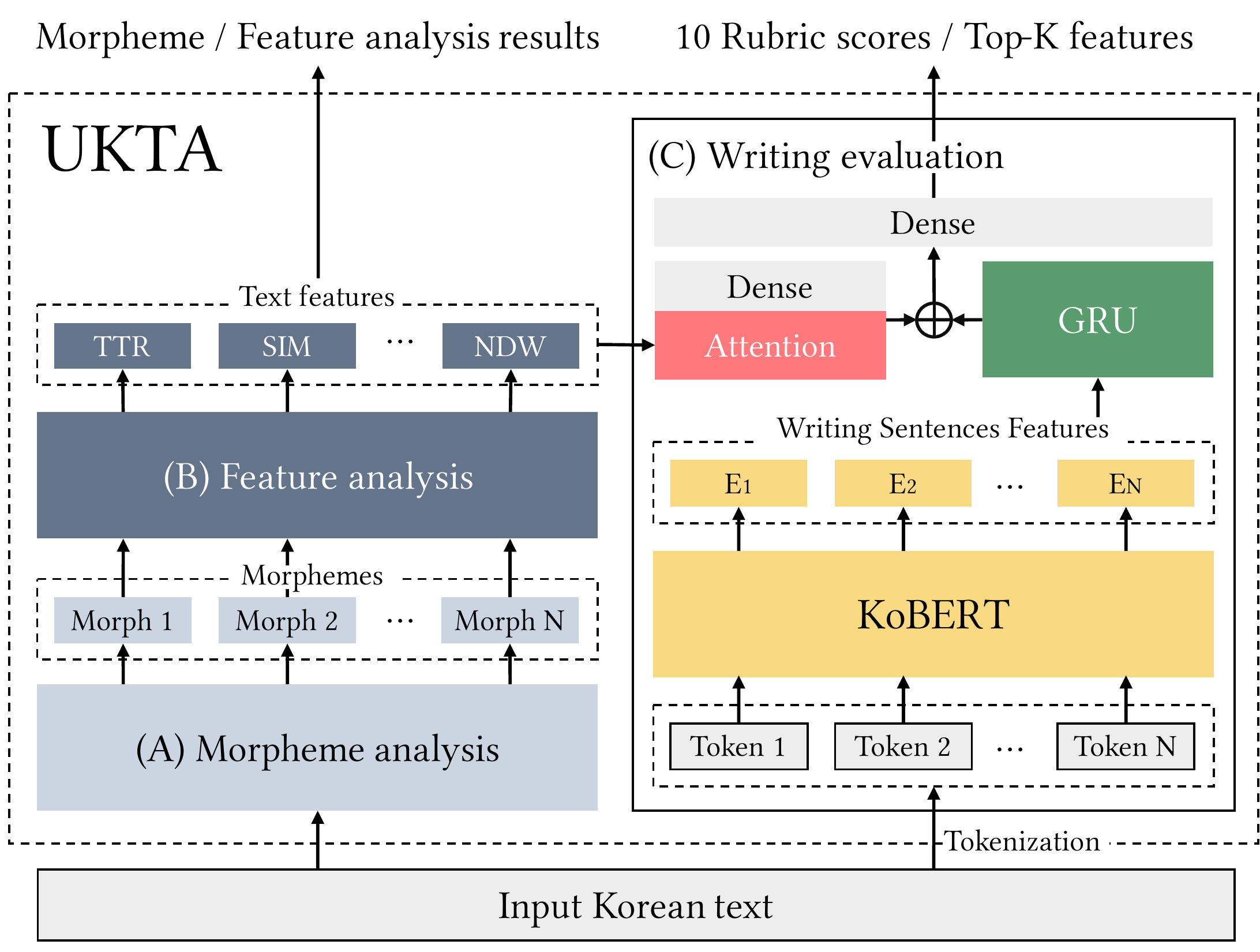}
\centering

\caption{Illustrative overview of \textsf{UKTA}.} 
\label{fig:approach}
\end{figure}

This section introduces \textsf{UKTA} (\textbf{U}nified \textbf{K}orean \textbf{T}ext \textbf{A}nalyzer), which sequentially performs low-level morpheme analysis (in Section \ref{sec:04_model_morpheme}), mid-level lexical feature analysis (in Section \ref{sec:04_model_feature}), and high-level automatic writing evaluation (in Section \ref{sec:04_model_scoring}).
The overall process for our tool is illustrated in Figure \ref{fig:approach}.

\begin{figure*}[t]
  \includegraphics[width=1\textwidth]{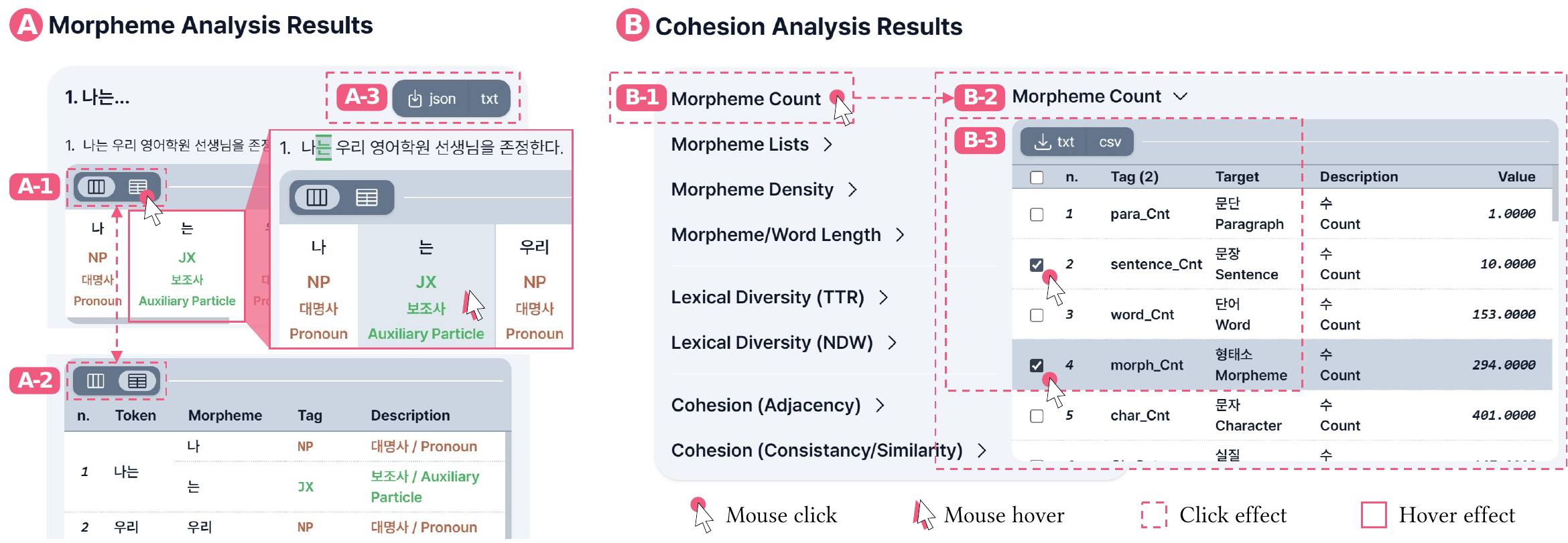}
  \centering
  \caption{\textsf{UKTA} functionality. \normalfont{\textbf{(A) Functionality in morpheme analysis results}: Providing both table (A-1) and list (A-1) format, with an interactive and intuitive intuitive interface; results can be downloaded in JSON and TXT formats (A-3). \textbf{(B) Functionality in lexical feature analysis results}: Provided as categorized lexical features (B-1) with a list format (B-2); results can be downloaded in TXT and CSV format with selected features (B-3).}}
  \label{fig:functions}
  \vspace{\baselineskip}
\end{figure*}

\subsection{Morpheme Analysis\label{sec:04_model_morpheme}}

This section describes the importance of Korean morpheme analysis and outlines the methods.
Morpheme analysis is the first step, low-level analysis before conducting Korean writing evaluation and lexical feature analysis. 
However, due to the nature of Korean, accurately segmenting wordpieces into morpheme units is challenging, as their forms can change due to different suffixes \cite{matteson2018rich}. 
Such errors can propagate to subsequent steps, including lexical feature analysis and writing evaluation. 

For example, morpheme analysis results should differ, `나\tiny{/NP}\normalsize{+는}\tiny{/JX}\normalsize{ (\textit{Na-Neun})' and `날}\tiny{/VV}\normalsize{+는}\tiny{/ETM}\normalsize{ (\textit{Nal-Neun})'}, even for the same type of wordpiece `나는 (\textit{Naneun})'. 
These errors can distort the values of some feature, reducing the reliability of feature analysis and writing evaluation. 
To minimize error propagation, we conducted morpheme analysis based on Bareun\footnote{https://bareun.ai/} analyzer, known for its highest accuracy, ensuring reliable results for subsequent lexical feature analysis and writing evaluation.

Our system provides morpheme analysis results in a clear, intuitive and user-friendly format as illustrated in Figure \ref{fig:functions}(A). It ensures easy interpretation and efficient analysis.

\subsection{Lexical Feature Analysis \label{sec:04_model_feature}}

This section introduces the process of mid-level feature analysis and describes various features.
After morpheme analysis, diverse features are numerically evaluated based on the morphemes. 
We provide numerical results for 294 features, broadly categorized into three groups: \textit{\textsf{basic lexical features}}, \textit{\textsf{lexical diversity}}, and \textit{\textsf{cohesion}}.
These features not only provide numerical information but also offer explainable insights for subsequent writing evaluation results.
Detailed descriptions of each group are provided below.

\subsubsection*{Basic Features} 
The basic lexical features of a text represent its fundamental linguistic composition. 
These features include measurements such as count, density, and length of morphemes or words. 
Accurate tagging and categorization of morphemes are essential for ensuring the precision of these metrics. 
Additionally, a list of sentences containing each morpheme is provided to clarify their contextual use Figure \ref{fig:teaser}(B-3). 
This detailed examination of basic features offers fundamental insights into the structural and linguistic properties of the text, serving as a basis for calculating more complex features.

\subsubsection*{Lexical Diversity}
Lexical diversity \cite{hout2007richness} 
 is measured based on the degree of connectivity between sentences or paragraphs, reflecting vocabulary depth and linguistic diversity \cite{ha2019lexical, kim2024korcat}.
This measure is evaluated through the calculation of lexical features such as Type-Token Ratio (TTR) and other diversity features. 
We provide each lexical diversity feature for all tokens, as well as for each specific morpheme. An example lexical diversity output is provided in Figure \ref{fig:teaser}(B-3).
A detailed description of the key feature for measuring lexical diversity included in our system is provided as follows.

\begin{itemize}[leftmargin=1.1em]
\item \textsf{\textit{Type-Token Ratios}}: TTR, RTTR (Root TTR), and CTTR (Corrected TTR) are fundamental features for calculating lexical diversity. 
These features tend to decrease as text length increases, making them suitable for comparisons between texts of similar length \cite{joan2013ttr}.
Formally, these features are calculated as:
\begin{equation}
    \text{TTR} = \frac{t}{w}, \quad
    \text{RTTR} = \frac{t}{\sqrt{w}}, \quad
    \text{CTTR} = \frac{t}{\sqrt{2w}}
\end{equation}
where \(t\) and \(w\) are the number of unique morphemes and the total number of morphemes in a given text, respectively.

\item \textsf{\textit{Equal segmented Type-Token Ratios}}: MSTTR (Mean Segmental TTR) and MATTR (Moving Average TTR) are both extensions of the traditional TTR, designed to address its sensitivity to text length. 
MSTTR calculates the average TTR over equal-length, and non-overlapping segments of a text, which standardizes the measure across different text lengths:
\begin{equation}
    \text{MSTTR} = \frac{1}{k} \sum_{i=1}^{k} \frac{t_i}{w_i} 
\end{equation}
where \(k = \left\lceil \frac{N}{n} \right\rceil\) is the number of non-overlapping segments, with \(N\) as the total number of tokens in the text and \(n\) as the window size. 
Here, \(t_i\) and \(w_i\) denote the number of unique tokens and the total number of morphemes in the \(i\)-th segment, respectively.
MATTR, in contrast, uses a moving window to compute TTR over overlapping segments, providing a more stable and consistent evaluation of lexical diversity across varying text lengths:
\begin{equation}
    \text{MATTR} = \frac{1}{k'} \sum_{i=1}^{k'} \frac{t_i}{w_i} = \frac{1}{k'n} \sum_{i=1}^{k'} t_i
\end{equation}
where \(k' = N-n+1\) is the number of overlapping segments.

\item \textsf{\textit{Textual lexical diversity}}:  
MTLD (Measure of Textual Lexical Diversity) addresses the sensitivity of TTR to text length by measuring how long a sequence of words must be to reach a fixed threshold of TTR decline. 
This approach provides a more stable and length-independent measure of lexical diversity.
In other words, MTLD is determined as the mean length of non-overlapping segments of varying length that satisfies the following:
\begin{equation}
    \text{MTLD} = \frac{N}{K}, \text{ where }\frac{t_i}{w_i} \leq \theta_\textrm{TTR}\text{ for all }1\leq i\leq K 
\end{equation}
\textit{i.e.}, \(K\) is the largest number of segments where the TTR value of each segment is below a predetermined threshold \(\theta_\textrm{TTR}\).

\item \textsf{\textit{Probabilistic lexical diversity}}: HD-D (Hypergeometric Distribution of Diversity) calculates the probability of encountering different morpheme types within randomly sampled subsets of the text, accounting for both morpheme occurrence and text length. 
Formally, HD-D calculates the average probability that each unique morpheme token \(t\) will appear at least once within a random sample of size \(S\):
\begin{equation}
    \text{HDD} = \frac{1}{S} \sum_{t} \left[ 1 - \frac{\binom{N - f_t}{S}}{\binom{N}{S}} \right]
\end{equation}
where \(f_t\) is the number of occurrences of token type \(t\) in the text. 

\item \textsf{\textit{Model-based lexical diversity}}:  
Voc-D estimates the relationship between tokens and types across various sample sizes, adjusting for the text length to provide a robust measure of vocabulary richness. This method addresses the limitations of raw TTR by accounting for variations in text length and complexity:
\begin{equation}
    \text{VOCD} = \argmin_{D} \sum_{n=35}^{50} \left( \overline{\text{TTR}}_n - \frac{D}{D+n} \right)^2
\end{equation}
where $\overline{\text{TTR}}_n$ is the empirical mean TTR of 100 random subsamples for size $n$, $D$ is the VOCD score that minimizes the difference between the empirical TTR values and the theoretical curve \(\frac{D}{D+n}\).
\end{itemize}

\subsubsection*{Cohesion}

Cohesion assesses topic consistency within a paragraph and the similarity of meanings between sentences \cite{kim2024korcat}.  
First, the topic sentence is identified by comparing the extracted keyword with each sentence in the paragraph (topic consistency), followed by calculating the similarity between the topic sentence and the remaining sentences (sentence similarity). 
\textsf{UKTA} uses KeyBert \cite{maarten2023keybert} for extracting key topics (keywords) and SBERT \cite{nils2019sbert} for measuring sentence similarity. 
Additionally, lexical overlap is used as a measure of cohesion, assessing the shared morphemes between adjacent sentences or paragraphs \cite{crossley2019taaco, kim2024korcat}. 
Two main types of lexical overlap are used: adjacent overlap, which counts the number of overlapping morphemes, and binary adjacent overlap, which only checks for their presence.
A higher lexical overlap indicates a stronger structural similarity in the text.

\subsection{Automatic Writing Evaluation\label{sec:04_model_scoring}} 

This section introduces a high-level automatic writing evaluation process that integrates the previously suggested low- and mid-level lexical features with the existing Korean automated writing evaluation model. 
The automatic writing evaluation task can be formally described as follows: 
Given an essay \( X = \{x_i\}_{i=1}^{N} \), consisting of \( N \) sentences, the objective is to predict 10 evaluation scores \( y_{i=1}^{10} \) corresponding to distinct evaluation criteria (commonly referred to as \textit{rubric}) such as grammar, vocabulary, and consistency. 

The architecture of our automatic writing evaluation model is shown in Figure \ref{fig:approach}(c). 
Previous Korean automated writing evaluation models have primarily focused on raw text, without considering the overall characteristics of the essay \cite{lee2022argument, lee2023pasta}. 
However, our model is capable of training on both the raw essay features (\textit{i.e.}, sentence-level features) and the overall characteristics (\textit{i.e.}, essay-level features), including basic lexical features, lexical diversity, and cohesion.
These essay-level features provide a comprehensive perspective of the essay’s quality. 
These features are derived from the Korean morpheme analyzer, enabling the model to perform accurate essay-level analysis during training. 
This process improves the overall accuracy of the writing evaluation.
Finally, we use the attention weights from the attention layer to emphasize the importance of different essay-level features for each sample, highlighting which features contributed to the model's predictions. 
Unlike previous approaches, we provide multiple-view analysis results using these attention scores, enhancing both the reliability and explainability of the writing evaluation results.

In summary, we utilize both i) sentence-level and ii) essay-level representations, then iii) combining them for a reliable and explainable writing evaluation.
The detailed description of the proposed model is as follows:

\begin{enumerate}[label={{\roman*)}}, leftmargin=1.1em]
\item {\textsf{\textit{Extracting sentence-level representations.} } 
We extract sentence-level representations of the essay using a pre-trained language model and a bidirectional Gated Recurrent Unit (BiGRU), as illustrated in the bottom of Figure \ref{fig:approach}(c).
The process begins by dividing the essay into $N$ sentences, and each sentence is tokenized.
The tokenized sentences are input into KoBERT\footnote{https://github.com/SKTBrain/KoBERT}, a BERT model pre-trained on a large-scale Korean corpus.
This step produces $N$ embedding vectors, denoted as $\mathbf{e}_1, \mathbf{e}_2, \dots, \mathbf{e}_N$.
The resulting sentence-level embedding vectors are subsequently fed into a BiGRU. 
These embedding vectors are then passed through a BiGRU, which computes the final sentence-level representation of the essay using its last hidden state vector, $\mathbf{h} = [\overrightarrow{\mathbf{h}}; \overleftarrow{\mathbf{h}}]$.
}
\item {\textsf{\textit{Extracting essay-level representations.} } 
We first extract the lexical features $\mathbf{f} \in \mathbb{R}^{294}$, which consist of 294 values from the raw text. 
These features are normalized using a standard scaler:
\begin{equation}
\mathbf{f'} = \frac{\mathbf{f} - \mu}{\sigma}
\end{equation}
where $\mu$ and $\sigma$ represent the mean and standard deviation calculated from the feature, respectively. 
This normalization process ensures that each feature is on the same scale and comparable. 
The normalized feature vector $\mathbf{f'}$ is then passed through an attention layer, which assigns different importance weights $\mathbf{A}$ to each feature. 
The attention-weighted vector is computed through an element-wise multiplication of the attention weights and the normalized features:
\begin{equation}
\mathbf{f_A} = \mathbf{A} \odot \mathbf{f'}
\end{equation}
where $\odot$ denotes element-wise multiplication. 
This operation emphasizes the most relevant features for the task. 
The attention-weighted vector $\mathbf{f_A}$ is then passed through a dense layer along with $\mathbf{f'}$, which outputs the final essay-level representation $\mathbf{v_e}$.
}
\item {\textsf{\textit{Combining sentence- and essay-level representations.} } 
Finally, the generated sentence-level representation vector $\mathbf{h}$ is concatenated with the essay-level representation vector $\mathbf{v_e}$. 
This combined vector is then passed through a linear layer followed by a sigmoid activation function to predict essay scores for 10 evaluation rubric criteria. 
During the training, the mean squared error (MSE) loss function is used.
}

\end{enumerate}

\section{Experiments}
\label{sec:05_experiments}
In this section, we validate the effectiveness of the proposed features in the automated writing evaluation system. 
To achieve this, we measure the performance of the automated writing evaluation model depicted in Figure \ref{fig:approach}, and compare it with the baseline model. 
Additionally, we examine the importance of each feature for sample data by analyzing the attention weights from the attention layer.

\subsection{Experimental Settings \label{sec:setting}}
{The purpose of our experiment is to verify whether the feature scores obtained from our text analysis system lead to performance improvements when incorporated into the training process of the automated writing evaluation model. Additionally, we aim to demonstrate the explainability of this automated evaluation tool by analyzing the predicted evaluation scores on sample data and assessing the importance of each feature.}

\subsubsection*{Dataset and Baseline model\label{sec:dataset}}
{For our experiment, we used the \textsf{Essay Evaluation Dataset} provided by \textsf{AI-HUB}. This dataset is the largest writing evaluation dataset available in Korea, consisting of essays written by students ranging from 4th grade of elementary school to 3rd grade of high school. It contains approximately 46,000 essays, which are categorized into five main types. The essays can be further divided into 50 topics, with around 900 essays per topic. A total of 46,000 essays were separated by topic, with approximately 40,000 used as training data and 6,000 used as test data. The essays are evaluated using a trait scoring system, with 10 distinct evaluation rubric scores assigned to each essay. The evaluation scores, used as labels, were assigned by human raters and reflect the assessments made by reliable experts in Korean writing evaluation.}

{To verify the effectiveness of the proposed features in the automated evaluation tools, we use the automated evaluation model provided by \textsf{AI-HUB} as the baseline. The baseline model utilizes only a PLM and a BiGRU. Sentence-level embedding vectors obtained from KoBERT are processed through the BiGRU to produce an essay representation vector, which is then passed through a single linear layer to output scores for 10 evaluation criteria.}

\subsubsection*{Evaluation metrics\label{sec:metric}}
We utilized both accuracy and Quadratic Weighted Kappa (QWK) as the evaluation metrics to evaluate the model performance. Accuracy is a ratio of the number of essays with correctly predicted scores to the total number of essays. QWK measures the agreement between two raters to reflect their consistent ratings. The formula for calculating QWK is as follows:
\begin{equation}
\text{QWK} = 1 - \frac{\sum_{i,j} w_{i,j} O_{i,j}}{\sum_{i,j} w_{i,j} E_{i,j}}
\end{equation}
where \( O_{i,j} \) represents the observed frequency matrix, \( E_{i,j} \) is the expected frequency matrix, and \( w_{i,j} \) is the weight assigned based on the squared difference between the scores \( i \) and \( j \).

\begin{table}[] 
\caption{Quantitative evaluation between baseline and \textsf{UKTA} by rubric scores. \normalfont{The best performance is marked in \textbf{bold}.}}
\centering
\setlength\tabcolsep{2.5pt}
\resizebox{1\columnwidth}{!}{
\begin{tabular}{clcccc}
\toprule
\multirow{2}{*}{\textbf{Major-rubric}} & \multirow{2}{*}{\textbf{Rubric}} & \multicolumn{2}{c}{\textbf{Accuracy}} & \multicolumn{2}{c}{\textbf{QWK}}   \\ \cmidrule{3-6} 
                              &                         & Baseline   & \textsf{UKTA}  & Baseline & \textsf{UKTA} \\ 
\cmidrule(r){1-2}\cmidrule{3-6}
\multirow{3}{*}{Expression}   & Grammar                 & 0.601      & 0.601           & 0.280    & 0.280          \\
                              & Vocabulary                   & 0.638      & \textbf{0.643}  & 0.337    & \textbf{0.343} \\
                              & Sentence Expression                & 0.713      & \textbf{0.715}  & 0.875    & \textbf{0.883} \\ 
\cmidrule(r){1-2}\cmidrule{3-6}
\multirow{4}{*}{Organization} & Inter-paragraph Structure         & 0.540      & \textbf{0.544}  & 0.347    & \textbf{0.357} \\
                              & In-paragraph Structure            & 0.748      & \textbf{0.753}  & 0.896    & \textbf{0.900} \\
                              & Structure Consistency             & 0.677      & \textbf{0.682}  & 0.857    & \textbf{0.861} \\
                              & Length                  & 0.725      & \textbf{0.748}  & 0.643    & \textbf{0.727} \\ 
\cmidrule(r){1-2}\cmidrule{3-6}
\multirow{3}{*}{Content}      & Topic Clarity                 & 0.636      & \textbf{0.641}  & 0.361    & \textbf{0.370} \\
                              & Originality                 & 0.615      & \textbf{0.621}  & 0.069    & \textbf{0.172} \\
                              & Narrative            & 0.599      & \textbf{0.622}  & 0.421    & \textbf{0.488} \\ \midrule
\textbf{Average} && 0.649      & \textbf{0.657}  & 0.509    & \textbf{0.538} \\ 
\bottomrule

\end{tabular}
}
\label{tb:Comparison}
\end{table}

\begin{table*}[th!]
\centering
\caption{\textbf{Qualitative evaluation between Korean writing evaluation low score (first row) and high score (second row), along with their Top-10 feature analysis.} \textnormal{The source of the translated input text, with the original typos properly reflected. The original low scoring text is in Figure \ref{fig:teaser}(A). The background colors of Top-10 feature correspond to \textred{\textit{\textsf{basic lexical features}}}, \textblue{\textit{\textsf{lexical diversity}}}, and \textpurple{\textit{\textsf{cohesion}}}, respectively. The tags for the Top-10 features used are provided in the table notes.
}}
\resizebox{1\textwidth}{!}{%
\begin{threeparttable}
\setlength\tabcolsep{5pt}
\begin{tabular}{p{0.75\textwidth} lcrlr}
    \toprule
    \textbf{Input text (Translated)} & \multicolumn{3}{c}{\textbf{Rubric score}}  & \multicolumn{2}{c}{\textbf{Top-10 feature}} \\
    \cmidrule(){1-1} 
    \cmidrule(lr){2-4}
    \cmidrule(){5-6}
    \multirow{5}{=}{\justifying I respct my Englishacademy teacher. I met teacher about 1-2 years ago, and the meeting was bit special. I got fired from my previous English academy due to some incident, and with an anger, I came to this academy where my friend was going. It might not seem a big deal, but no. The reason I respect this teacher is that even though I honestly think I’m a bit troublesome, teacher hendles not only me but also other friends who are even more trouble, and halps us stay focosed on study. Also, this place isn’t just an English academy, I feel I’m learning a lot of important skills for life, manners, and social skills, so I’m very satisfied. Also my grades improved a lot too. Honestly, colling it a life academy wouldn’t be too much. After graduatingfromhere, I definitely wanttoget into Dongguk Univ's PE department, visit teacher sometimes, and show myself to juniors, like `I succeeded lkie this, so you guys can do it to'. I want to improve my skills while attending this academy, get a good university, and become a pride for my teacher.
    }
    & \textbf{Grammar} & 2 && \blue{\textbf{\texttt{FL\_MSTTR}}} & 0.58 \\
    & \textbf{Vocabulary} & 3 && \red{\textbf{\texttt{CL\_Den}}} & 0.50 \\
    & \textbf{Sentence} & 1 && \purple{\textbf{\texttt{ASO\_ALN}}} & 9.00 \\
    & \textbf{In-paragraph} & 2 && \blue{\textbf{\texttt{NNB\_MSTTR}}} & 0.50 \\
    & \textbf{Inter-paragraph} & 3 && \red{\textbf{\texttt{NNL\_Den}}} & 0.19 \\
    & \textbf{Consistency} & 2 && \red{\textbf{\texttt{NNCL\_Den}}} & 0.37 \\
    & \textbf{Length} & 3 && \blue{\texttt{\textbf{VV\_RTTR}}} & 4.70 \\
    & \textbf{Clarity} & 2 && \purple{\textbf{\texttt{AvgSenSimilarity}}} & 0.26 \\
    & \textbf{Originality} & 2 && \red{\textbf{VCL\_Den}} & 0.41 \\
    & \textbf{Narrative} & 2 && \blue{\textbf{\texttt{IC\_RTTR}}} & 1.00\\
    \cmidrule(){1-1}  
    \cmidrule(lr){2-4}
    \cmidrule(){5-6}
     \multirow{5}{=}{\justifying A South Korean singer-songwriter. A member of the mixed-gender group Jaurim, where she serves as the vocalist. In 2004, she received the Special Jury Award at the Mnet Km Music Video Festival and the Beautiful Lyrics Award at the KBS Correct Language Awards. In 2011, she was selected as the Musician of the Year by netizens at the 8th Korean Music Awards. Even now, with her 50s just around the corner, she is famous for looking much younger. Although she has naturally good skin, in her own words, because she takes care of her skin very thoroughly on a regular basis. Among Jaurim fans, there's a running joke that she divides her age by 1/4 and gives the rest to the other members. On broadcasts, she looks incredibly charismatic and seems strong-willed, but in reality, she is surprisingly slim and pretty. She has also been a model for a cosmetic brand. She was raised with such strict discipline by her father that it felt stifling, and even now, she places great importance on manners ... (omitted).
    }
    & \textbf{Grammar} & 3 && \purple{\textbf{\texttt{ASO\_CLN}}} & 23.00 \\
    & \textbf{Vocabulary} & 3 && \blue{\textbf{\texttt{NNP\_NDW}}} & 14.00 \\
    & \textbf{Sentence} & 3 && \blue{\textbf{\texttt{NNG\_NDW}}} & 134.00 \\
    & \textbf{In-paragraph} & 3 && \red{\textbf{\texttt{FL\_Den}}} & 0.42 \\
    & \textbf{Inter-paragraph} & 3 && \red{\textbf{\texttt{XFL\_Den}}} & 0.04 \\
    & \textbf{Consistency} & 3 && \blue{\textbf{\texttt{VV\_RTTR}}} & 5.81 \\
    & \textbf{Length} & 3 && \blue{\textbf{\texttt{MM\_VOCDD}}} & 0.00 \\
    & \textbf{Clarity} & 3 && \blue{\textbf{\texttt{NNB\_MSTTR}}} & 0.52 \\
    & \textbf{Originality} & 2 && \blue{\textbf{\texttt{E\_NDW}}} & 156.00 \\
    & \textbf{Narrative} & 3 && \purple{\textbf{\texttt{ASO\_FLN}}} & 23.00\\
    \bottomrule
\end{tabular}
\begin{tablenotes}
\item[-] Morpheme tag: \texttt{CL} (Content lemmas), \texttt{DD} (Determiner), \texttt{E} (Ending), \texttt{FL} (Function lemmas), \texttt{IC} (Interjection), \texttt{NNB} (Dependent noun), \texttt{NNG} (Common noun), \texttt{NNL} (Content nown), \texttt{NNP} (Proper Noun), \texttt{VCL} (Content coupla), \texttt{VV} (Verb), \texttt{XFL} (Affix formal) 
\item[-] Lexical feature tag: \texttt{Den} (Density), \texttt{MSTTR} (Mean Segmental
TTR), \texttt{RTTR} (Root TTR), \texttt{NDW} (Number of Different Words), \texttt{VOCDD} (Vocd-D)
\item[-] Cohesion tag: \texttt{ASO} (Adjacent Sentence Overlap), \texttt{ALN} (All Lemmas Normed), \texttt{CLN} (Content Lemmas Normed), \texttt{FLN} (Function Lemmas Normed)
\end{tablenotes}
\end{threeparttable}
}
\label{tb:feature}
\end{table*}

\subsubsection*{Implementation details\label{sec:imp_detail}}
{The experiments were conducted in an environment equipped with an Intel(R) Core(TM) i7-13700F and an NVIDIA RTX 4090. All hyperparameters were kept constant throughout the experiments: dropout is set to 0.5, learning rate to 0.001, and the number of epochs to 100, respectively. Early stopping was employed to save only the best-performing model. The activation function of the final linear layer is the sigmoid function. 
Each experiment (\textit{i.e.}, baseline and \textsf{UKTA}) was performed five times, and the average values were used as a final result.
}

\subsection{Experimental Results\label{sec:result}}

\subsubsection*{Quantitative results}

Table \ref{tb:Comparison} presents the quantitative results of the baseline automated writing evaluation model for each evaluation metric, along with the performance of the \textsf{UKTA}. 
Both the accuracy and QWK metrics show a significant improvement when feature scores are incorporated. A closer examination of each rubric reveals that 9 out of the 10 rubrics show improvements in both accuracy and QWK, indicating that the model benefits from the feature scores in evaluating the appropriateness of "expression", "organization", and "content" in the essays. Specifically, in the "expression" category, there is a notable increase in performance for "vocabulary" and "sentence expression"; in the "organization" category, "length" shows improvement; and in the "content" category, "originality" and "narrative" demonstrate significant gains. These findings suggest that the feature scores most effectively help the model evaluate the appropriateness of vocabulary, sentence expression, adequacy of length, narrative quality, and originality of content. Finally, when averaging the performance across all rubrics, the accuracy increased from 0.649 to 0.657, and the QWK improved from 0.509 to 0.538.

\subsubsection*{Qualitative results}
We also select an essay from each of the highest-scoring and lowest-scoring groups within the same topic, and consequently analyzed the top-10 lexical features that the model emphasized for each group.  
The detailed contents of the essays are presented in Table \ref{tb:feature}, and the qualitative evaluation results for each group are as follows:

\begin{itemize}[leftmargin=1.1em, ]
    \item \textsf{\textit{Low-scoring essay.}} In the low-scoring essay, lexical features related to content morphemes (\textit{i.e.,} \texttt{CL\_Den}, \texttt{NNB\_MSTTR}, \texttt{NNL\_Den}, \texttt{NNCL\_Den}, \texttt{VV\_RTTR}, and \texttt{VCL\_Den}) accounted for 60\% of the model's influence. In contrast, lexical features related to grammatical morphemes (\textit{i.e.,} \texttt{FL\_MSTTR}) had minimal impact, contributing only about 10\%. 
    Grammatical morphemes are essential for determining subjects, objects, and predicates, as well as differentiating subtle meanings in expressions
    The simplicity of the sentence structure likely reduced the role of grammatical morphemes, leading to less weight being assigned to features related to grammatical morphemes in the evaluation. 
    Additionally, cohesion-related features contributed approximately 20\% to the score, reflecting mid-level performance in "content" and "organization".
    Notably, interjections, which are uncommon in formal Korean writing, influenced the score by 10\%. 
    This was likely due to spelling errors misinterpreted by the morpheme analyzer. These findings highlight the potential impact of misanalysis or outliers on quality predictions.

    \item \textsf{\textit{High-scoring essay.}} In the high-scoring essay, lexical features related to content morphemes (\textit{i.e.,} \texttt{NNP\_NDW}, \texttt{NNG\_NDW}, \texttt{VV\_RTTR}, \texttt{MM\_VOCDD}, and \texttt{NNB\_MSTTR}) and grammatical morphemes (\textit{i.e.,} \texttt{FL\_Den}, \texttt{XFL\_Den}, and \texttt{E\_NDW}) play a more balanced role. 
    This suggests that increased semantic complexity resulted in greater weight being assigned to features related to grammatical morphemes. 
    Among these, ending diversity and affix density had a particularly strong influence, contributing to sentence complexity and nuance.
    Content morphemes, especially proper nouns and dependent nouns, also had a significant impact.
    The more closely the essay aligned with the "My Hero" topic, which involves narrating a person's story, the greater the impact of these morphemes. This indicates that morpheme diversity analysis can provide insights into how well the content adheres to the topic. The influence of cohesion in this text is 20\%, with the overlap of content lemmas showing the greatest impact among all features. It suggests that a well-organized, cohesive essay contributes to higher scores.
\end{itemize}
In short, by comparing the features of both high-scoring and low-scoring essays, we demonstrated that the model's key influential features—content morphemes, grammatical morphemes, and cohesion—can be qualitatively analyzed, 
providing meaningful insights into essay quality.

\section{Conclusion}
\label{sec:06_conclusion}

In this work, we propose \textsf{UKTA}, a comprehensive Korean text analysis and writing evaluation system designed for practical use. Unlike existing Korean writing evaluation tools, \textsf{UKTA} provides an automated writing evaluation score along with analyses such as morpheme analysis, lexical diversity features, and cohesion, enhancing the evaluation explainability and reliability. Additionally, our proposed evaluation model based on lexical features outperforms baselines relying solely on raw text data. By analyzing the importance of each feature, \textsf{UKTA} demonstrates its ability to identify the factors influencing Korean writing evaluation.

\textsf{UKTA} opens new possibilities for Korean writing evaluation, offering educators transparent scoring metrics and researchers deeper insights into language features. Future studies could explore its application in diverse educational settings or further refine its feature set to enhance accuracy and adaptability. With its comprehensive approach, \textsf{UKTA} is poised to become a reliable tool for both academic research and practical writing assessment.

\section*{Acknowledgements}
This work was supported in part by the National Research Foundation of Korea (NRF) grant funded by the Korea government (MSIT) (No. NRF-2022R1C1C1012408), in part by Institute of Information \& communications Technology Planning \& Evaluation (IITP) grants funded by the Korea government (MSIT) (No. 2022-0-00448/RS-2022-II220448, Deep Total Recall: Continual Learning for Human-Like Recall of Artificial Neural Networks, and No. RS-2022-00155915, Artificial Intelligence Convergence Innovation Human Resources Development (Inha University)), and in part by the Ministry of Education of the Republic of Korea and the National Research Foundation of Korea (No. NRF-2023S1A5A2A21085373).

\balance
\bibliographystyle{ACM-Reference-Format}
\bibliography{main.bib}

\end{document}